%% file: main.tex
\documentclass{article} 
\usepackage{iclr2027_conference,times}

\usepackage{latexsym}
\usepackage[T1]{fontenc}
\usepackage[utf8]{inputenc}
\usepackage{microtype}
\usepackage{graphicx}
\usepackage{amsmath, amssymb, amsthm}
\usepackage{booktabs}
\usepackage{enumitem}
\usepackage{xcolor}
\usepackage{hyperref}
\usepackage{url}
\usepackage{tikz}
\usetikzlibrary{arrows.meta, positioning, fit, calc}
\usepackage{tabularx}
\usepackage{siunitx}
\title{CoLMbo-SV: A Grounded Language Model for Explainable Speaker Verification}
\author{Massa Baali, Sarthak Bisht, Ziyue Qiu, Joseph Konan, Rita Singh, Bhiksha Raj \\
Carnegie Mellon University, USA \\
\texttt{mbaali@cs.cmu.edu}
}
\iclrfinalcopy 

\begin{document}
\maketitle
\lhead{Preprint. Under review.}   

\begin{abstract}
\input{sections/abstract}

\end{abstract}

\section{Introduction}
\input{sections/intro}
\section{Related work}
\input{sections/related}

\section{CoLMbo-SV}
\input{sections/method}
\section{VoxReason}
\label{sec:dataset}
\input{sections/dataset}
\section{Evaluating grounded reports}
\input{sections/eval}
\section{Results}
\input{sections/results}
\section{Ablations}
\input{sections/ablation}
\section{Does the report explain the decision?}
\input{sections/tcav}

\section{Conclusion}
In this work, we introduced CoLMbo-SV, a speaker language model that compares two recordings and explains the comparison in a structured report grounded in acoustic measurements. To train it, we built VoxReason, a dataset of paired recordings with measured acoustic properties and validated comparison reports. CoLMbo-SV reaches 0.99\% EER on VoxCeleb1-O, well ahead of audio-language models fine-tuned on the same data, and 82\% of the numbers in its reports match the measurements even though no number appears in any baseline prompt. We also found that the way measurements reach the language model matters: given as text, the model cites them exactly, while given as learned soft tokens, it mostly does not. We then asked a harder question: does a correct report actually explain the decision? By separating what the speaker embedding encodes, what the decision is sensitive to, and what the report discusses, we found that it does not yet. The reports are accurate, but they discuss largely the same features for every pair, regardless of which evidence drove that pair's verdict, and telling the model which evidence mattered does not change this. A report can be factually correct and still fail to explain the decision. We hope the evaluation framework introduced here helps future work close this gap, moving from reports that describe the recordings to reports that explain the verdict.

\paragraph{Limitations.}
Our evaluation covers read and interview speech (TIMIT, VoxCeleb, SITW);
telephone channels, forensic field recordings and very short comparisons
are natural next targets. The decision-relevant set is estimated by a
sensitivity analysis over one system's two decision paths; interventional
tests, sex-stratified probes and semantic rather than lexical matching of
feature mentions would sharpen it. Finally, CoLMbo-SV produces reports that
are grounded but not yet decision-selective. Because prompting does not
close this gap, we see supervision that ties each report to the concepts
driving that pair's verdict, which our framework now makes measurable, as
the key step toward reports that explain the decision.

\subsection*{AI use statement}

Generative AI tools were used in two ways. First, as part of the method:
the VoxReason comparison reports are generated by an LLM from measured
acoustic values and filtered by a deterministic numerical validator and an
LLM-as-judge (\S\ref{sec:dataset}). Second, as research assistance:
exploring experimental directions, and polishing and reviewing the
manuscript for clarity and consistency. All research ideas, experimental design, results and
conclusions are the authors' own; the authors reviewed and verified all
AI-assisted text.

\subsection*{Ethics statement}

Speaker verification has both legitimate and potentially harmful applications. Legitimate use includes secure authentication, forensic
phonetic analysis under court-admissible standards, and accessibility applications. Potential harms include surveillance of vulnerable
populations and over-reliance on machine output in legal proceedings. The structured analysis trace this work introduces is intended to
support the legitimate forensic use case by making the model's
reasoning auditable. We caution against deploying the system in legal contexts without the trace explicitly inspected by qualified phoneticians, and against use cases for which speaker verification was not designed.

\subsection*{Reproducibility statement}

VoxReason's construction is specified in \S\ref{sec:dataset} and
Appendix~\ref{app:voxreason} (sources, pair stratification, measurement
set, generator, validator tolerances and yield); the architecture, losses,
token weights, stage schedule and decoding settings in \S\ref{sec:method};
the trial lists, the 600-pair evaluation set, the probe, sensitivity and
decision-grounding procedures in \S\ref{sec:eval} and
Appendix~\ref{app:tcav}; and every ablation configuration in
\S\ref{sec:ablation} and Appendix~\ref{app:decoding}. The speaker encoder
and the language-model backbone are public checkpoints.

\bibliography{references}
\bibliographystyle{iclr2027_conference}

\appendix
\input{sections/appendix}

\end{document}

%% file: sections/abstract.tex
Speaker verification systems achieve high accuracy but provide little account of the acoustic evidence behind their judgments. Making these systems inspectable requires exposing interpretable evidence while retaining the richer information on which their decisions depend. We present \textbf{CoLMbo-SV}, a speaker language model that combines strong speaker discrimination with structured, acoustically grounded comparison reports. By connecting a pretrained speaker encoder to a language model and supplying explicit acoustic measurements, CoLMbo-SV makes voice comparisons inspectable without restricting verification to the evidence verbalized in its reports. We additionally introduce \textbf{VoxReason}, paired recordings with measured acoustic properties and comparison reports filtered through numerical and qualitative checks, providing supervision for this combined capability. We also develop an evaluation framework that separates what acoustic information a speaker representation encodes, what influences the verification score, and what the generated report discusses. On VoxCeleb1-O, CoLMbo-SV achieves 0.99\% EER, reducing verification error by approximately 80\% relative to the strongest audio-language baseline fine-tuned on VoxReason, while attaining a numerical-grounding score of 0.82. Our analysis further demonstrates that acoustic correctness and decision relevance are distinct properties of an explanation, exposing a gap that numerical-grounding metrics miss. Together, these contributions substantially advance audio-language speaker verification, bring its accuracy toward that of dedicated speaker encoders while adding checkable acoustic reporting, and establish an empirical framework for connecting natural-language explanations to the decisions they explain.

%% file: sections/intro.tex
While speaker verification systems have become highly accurate, they provide no explanation of how they arrived at their decisions. Modern systems map recordings into speaker embeddings and compare them using a similarity score or a learned decision function \citep{snyder2018xvectors,desplanques2020ecapa,chen2022wavlm}. These representations capture information that distinguishes speakers, but the resulting score does not tell a user which acoustic similarities support a match, which differences count against it, or how recording conditions affect the comparison. Making that information accessible is the central challenge addressed in this paper. We want to be able to inspect the evidence available to a strong verifier while retaining the expressive representation that makes it effective.

An explanation need not reconstruct the \textit{entire} computation to be useful. A speaker representation may exploit information beyond the acoustic properties we can currently name or measure. Restricting verification to inferences achievable from a prescribed set of human-readable features would make those features a bottleneck on the decision. We take an alternate route --  \textit{post-hoc} explanations, which attempt to explain a performed computation \textit{after the fact}. Post-hoc explanations allow the verifier to use its full representation while exposing selected, interpretable aspects of the evidence. Such an account can be partial, e.g. it may explain the role of voice quality or pitch variability without claiming that these exhaust the basis of the decision. The requirement is that what it does explain is faithful. Generating a verbal rationale, whether before or after an answer, does not by itself establish this connection \citep{turpin2023language,jacovi2020faithfulness}.

This requirement separates two questions that a fluent report can easily be obscure. \emph{Does the report describe the recordings correctly?} And \emph{does it identify evidence relevant to the model's decision?} A statement that two voices have similar pitch can be acoustically correct even if pitch similarity played little role in their verification score. Conversely, the score may depend on information that the report never mentions. Recovering a property from a speaker embedding, showing that the decision is sensitive to that property, and explaining its role in a particular comparison are therefore separate and distinct achievements. Progress toward explainable verification requires a system in which these connections can be examined separately.
We introduce \textbf{CoLMbo-SV}, a speaker language model that combines strong verification with acoustic comparison reports and an audit of their relationship to the decision. CoLMbo-SV couples a pretrained speaker encoder to a language model through a learned mapper, allowing verification and report generation to draw on the speaker representation. Extracted acoustic measurements provide an explicit basis for numerical reporting, complementing the identity information carried by the learned embeddings. The system thus supports inspectable acoustic comparisons without restricting verification to the evidence verbalized in its reports.

To support this task, we construct \textbf{VoxReason}: paired recordings from TIMIT and VoxCeleb2 accompanied by acoustic measurements and structured comparison reports. The reports address recording conditions, prosody, voice quality, and segmental properties. We generate supervision from the measured evidence and filter candidate reports using numerical checks and an LLM judge. The resulting data provide a common basis for learning comparative reporting and evaluating whether its numerical claims agree with the extracted measurements. We distinguish this \emph{numerical grounding} from \emph{decision faithfulness}: correct measurements are necessary for a factual acoustic report, but are insufficient to explain a verification outcome.

CoLMbo-SV demonstrates that strong speaker verification can coexist with checkable acoustic reporting. With extracted measurements supplied as text, it achieves 0.99\% equal error rate on VoxCeleb1-O, while 82\% of the numerical claims in its reports match the extracted measurements under our evaluation criterion. Its verification error is substantially lower than that of the audio language models fine-tuned on VoxReason, approaching that of dedicated speaker encoders. These results establish a system that can both discriminate speakers and report acoustic evidence, allowing us to examine the further requirement for explanation: whether the evidence it reports is relevant to the decision it makes.

Accurate acoustic reporting does not by itself establish that a system explains its decisions. We therefore separate three questions: what acoustic information the speaker embedding encodes, which of these properties the verification score is sensitive to, and which the report discusses. Using speaker-disjoint concept probes and a sensitivity analysis based on concept activation vectors \citep{kim2018tcav}, we test whether reports preferentially discuss influential concept directions associated with the current pair rather than another pair. The results expose a gap that numerical grounding alone misses: CoLMbo-SV's reports describe acoustic evidence accurately but show little selectivity for the evidence influencing the particular decision. This distinction makes the central challenge of post-hoc explanation measurable: establishing that a report explains the decision, beyond correctly describing its inputs.

\noindent Our contributions are threefold:
\begin{itemize} [leftmargin=*]
    \item \textbf{CoLMbo-SV, a speaker language model combining strong verification with grounded acoustic reporting.} We develop a system that produces speaker-verification decisions alongside structured, numerically checkable comparisons. On VoxCeleb1-O, it substantially outperforms the audio language models fine-tuned on VoxReason, approaching the verification accuracy of dedicated speaker encoders while providing a natural-language account of the acoustic evidence.

    \item \textbf{VoxReason, a dataset for grounded comparative speaker analysis.} We construct paired-recording supervision that brings together speaker identity, measured acoustic properties, and structured comparison reports. Numerical validation and qualitative filtering support a resource for training speaker language models and evaluating their reports against explicit acoustic evidence.

    \item \textbf{A framework for evaluating whether acoustic reports explain verification decisions.} We connect acoustic-concept probing, decision-sensitivity analysis, and pair-specific report evaluation to distinguish what a representation encodes, what influences its output, and what its explanation says. This framework exposes a gap that numerical grounding and feature coverage miss: accurate acoustic reporting does not ensure that a report identifies the evidence influencing the particular decision.
\end{itemize}

%% file: sections/related.tex
\label{sec:related}
Post-hoc explanation is a large field. It includes feature attribution
\citep{lundberg2017shap,sundararajan2017ig}, concept-based analysis
\citep{kim2018tcav,koh2020concept} and prototypes \citep{chen2019prototype}.
For language models, it includes verbal rationales, whose faithfulness is
itself contested \citep{jacovi2020faithfulness,turpin2023language}.
These methods explain a computation already performed, in terms of the
model's own inputs or internal units. Speech poses a different problem.
Here an explanation must be expressed in a standard set of physically or
spectrally meaningful quantities, such as pitch, formants and voice quality,
that can be checked against the signal. To our knowledge, no post-hoc
explanation work addresses this setting or targets speech. Interpretable
speaker recognition instead builds interpretability into the verifier
\citep{benamor2022balr,ma2025expo}. Appendix~\ref{app:related} reviews
explainable speaker verification and post-hoc explanation in more detail.

%% file: sections/method.tex
\label{sec:method}

\begin{figure}[t]
    \centering
    \includegraphics[width=0.99\linewidth]{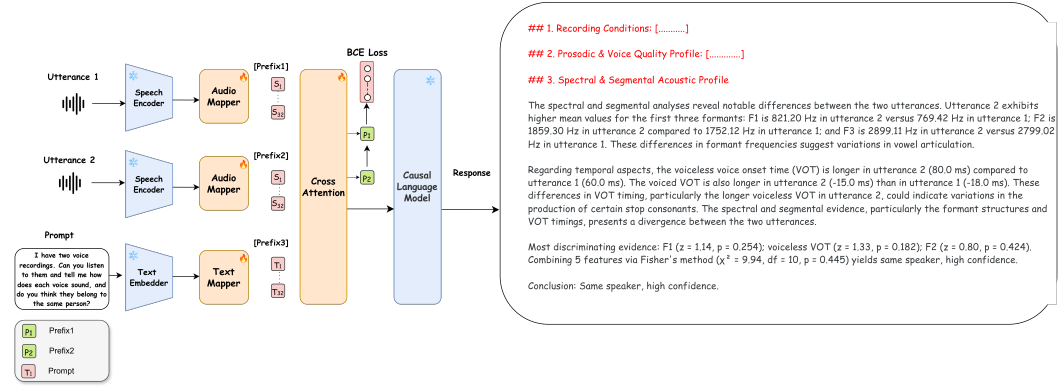}
    \caption{\textbf{CoLMbo-SV.} A frozen speaker encoder embeds each
    recording; a mapper turns the embeddings and the prompt into a
    soft-token prefix for a LoRA-adapted language model; a raw-evidence
    bypass carries the uncompressed embeddings to a verification head and
    to a comparison token in the prefix; the extracted measurements enter
    as text. The language model writes the report and the verdict scored
    for EER; the head is a second, continuous decision path.}
    \label{fig:sys_pipeline}
\end{figure}

CoLMbo-SV is a speaker language model for voice comparison. It takes two
recordings and a text prompt. It returns a structured report that compares
the two voices, cites acoustic measurements, and ends in a same- or
different-speaker verdict. A frozen speaker encoder embeds each recording.
A signal-processing front end extracts eighteen acoustic measurements from
each recording. A language model reads both and writes the report
(Figure~\ref{fig:sys_pipeline}). The design has one goal: verification
should keep the full speaker representation, while the report cites exact
values that can be checked against the signal.

To meet it, CoLMbo-SV carries three kinds of information by three separate
paths, so that none is a bottleneck on the others. The full speaker
representation reaches the decision uncompressed through a
\emph{raw-evidence bypass}. A lossy but expressive \emph{mapper} gives
the language model a prefix to write from. The \emph{measurements} reach
the language model as text, so the report can cite them exactly.

\paragraph{Speaker encoder.}
We use a W2V-BERT~2.0 backbone pretrained for speaker verification with a
curriculum ranking loss \citep{baali2026curr}, which yields a pooled speaker
embedding $\mathbf{e}\in\mathbb{R}^{192}$ per utterance. The encoder is
frozen throughout and identical in every result. It is deliberately not
constrained to the eighteen measured concepts: verification stays free to
use speaker information we cannot name.

\paragraph{Mapper.}
A transformer mapper \citep{mokady2021clipcap} projects each embedding into
$K{=}32$ soft tokens $\mathbf{S}_a,\mathbf{S}_b$ in the language model's
embedding space, a third projector maps the embedded prompt to
$\mathbf{S}_p$, and a four-layer cross-projection transformer mixes
$[\mathbf{S}_a;\mathbf{s}_{\text{sep}};\mathbf{S}_b;\mathbf{c};\mathbf{S}_p]$
into the prefix $\tilde{\mathbf{X}}$; $\mathbf{c}$ is the comparison
token defined next.

\paragraph{Raw-evidence bypass.}
Why not verify through the prefix? Because compression into soft tokens is
lossy for verification: the frozen encoder separates speakers almost
perfectly by cosine similarity, yet much of that evidence does not survive
the mapper. We therefore compute a raw comparison feature on the
unprojected, $\ell_2$-normalised embeddings,
$\mathbf{f}=[\hat{\mathbf{e}}_a;\hat{\mathbf{e}}_b;|\hat{\mathbf{e}}_a-\hat{\mathbf{e}}_b|;\hat{\mathbf{e}}_a\odot\hat{\mathbf{e}}_b;\cos(\hat{\mathbf{e}}_a,\hat{\mathbf{e}}_b)]$,
and route it around the mapper to two places. A \emph{verification head},
a two-layer MLP on $\mathbf{f}$ with a residual on the cosine,
$\ell=\mathrm{MLP}(\mathbf{f})+\gamma\cos(\hat{\mathbf{e}}_a,\hat{\mathbf{e}}_b)+\beta$,
is a near-perfect verifier from the first step. The same
feature is projected into the \emph{comparison token} $\mathbf{c}$ of the
prefix, so the written verdict can condition on evidence that never went
through the lossy projection; without this token the head separates the
speakers while the written verdict disagrees with it.

\paragraph{Measurement channel.}
The eighteen measurements of the two recordings are rendered as a compact
text block (name, value for utterance~1, value for utterance~2; about 400
tokens) appended to the prompt and read through the language model's own
embedding layer. We render them as text rather than compress them into
learned tokens on purpose: the aim is not only to transfer the information
but to preserve the exact values that make a report externally checkable.
The extractor is the one used to build VoxReason (\S\ref{sec:dataset}), so the system remains
audio-in, report-out; learned tokens are evaluated in \S\ref{sec:ablation}.

\paragraph{Language model and objective.}
$\tilde{\mathbf{X}}$ followed by the measurement block is the prefix of a
Gemma-4-E2B-it backbone with LoRA adapters \citep{hu2022lora} ($r{=}16$,
$\alpha{=}32$, attention and MLP projections, bfloat16). For a target report
$y_{1:T}$ the language-modelling loss is
$\mathcal{L}_{\text{LM}}=-\sum_t w_t\log P_\theta(y_t\mid\tilde{\mathbf{X}},y_{<t})/\sum_t w_t$
with prefix positions masked, and the head is trained with binary
cross-entropy $\mathcal{L}_{\text{BCE}}$ on $\sigma(\ell)$; the total is
$\lambda_{\text{LM}}\mathcal{L}_{\text{LM}}+\lambda_{\text{BCE}}\mathcal{L}_{\text{BCE}}$.
Because numerical claims and comparison directions are what the grounding
evaluation checks, $w_t{=}5$ on numeral tokens and on tokens expressing a
direction (``higher'', ``lower''), $w_t{=}1$ otherwise; the weights are used
in every configuration trained on evidence-block targets (stage~3 and
\S\ref{sec:ablation}).

\paragraph{Training as a curriculum.}
Three stages, each fine-tuned from the last, give the model one thing to
learn at a time. \emph{(1)~Learn to verify}: mapper, head and LoRA are
trained on report targets without an evidence block,
$\lambda_{\text{LM}}{=}\lambda_{\text{BCE}}{=}1$, 30k steps.
\emph{(2)~Learn to report without moving the verifier}: encoder and head
frozen, same targets, $\lambda_{\text{BCE}}{=}0.05$, 45k steps.
\emph{(3)~Connect the report to exact measurements}: targets that open
with a structured evidence block (value, comparison, difference, direction
per measurement), the measurement text block in the prompt and the weighted
loss, 30k steps. \S\ref{sec:ablation} shows what each stage buys.

\paragraph{Two decision paths.}
CoLMbo-SV therefore has two decision interfaces. The language model produces
the same/different conclusion of its report, and the \emph{verdict score}
derived from it (\S\ref{sec:eval}) is what every EER in this paper measures.
The verification head produces a continuous logit trained with the BCE
term and frozen after stage~1; it does not score the verdict, but it is a
second decision path over the same evidence, and \S\ref{sec:tcav} analyses
both because a report can be grounded without describing the evidence
either path uses. At inference the
measurements are extracted and rendered into the prompt and the language
model decodes greedily with a repetition penalty of 1.15 and 6-gram
blocking, up to 2{,}560 tokens; the verdict is parsed from the conclusion
line.

%% file: sections/dataset.tex
\begin{figure}[t]
    \centering
    \includegraphics[width=0.9\linewidth]{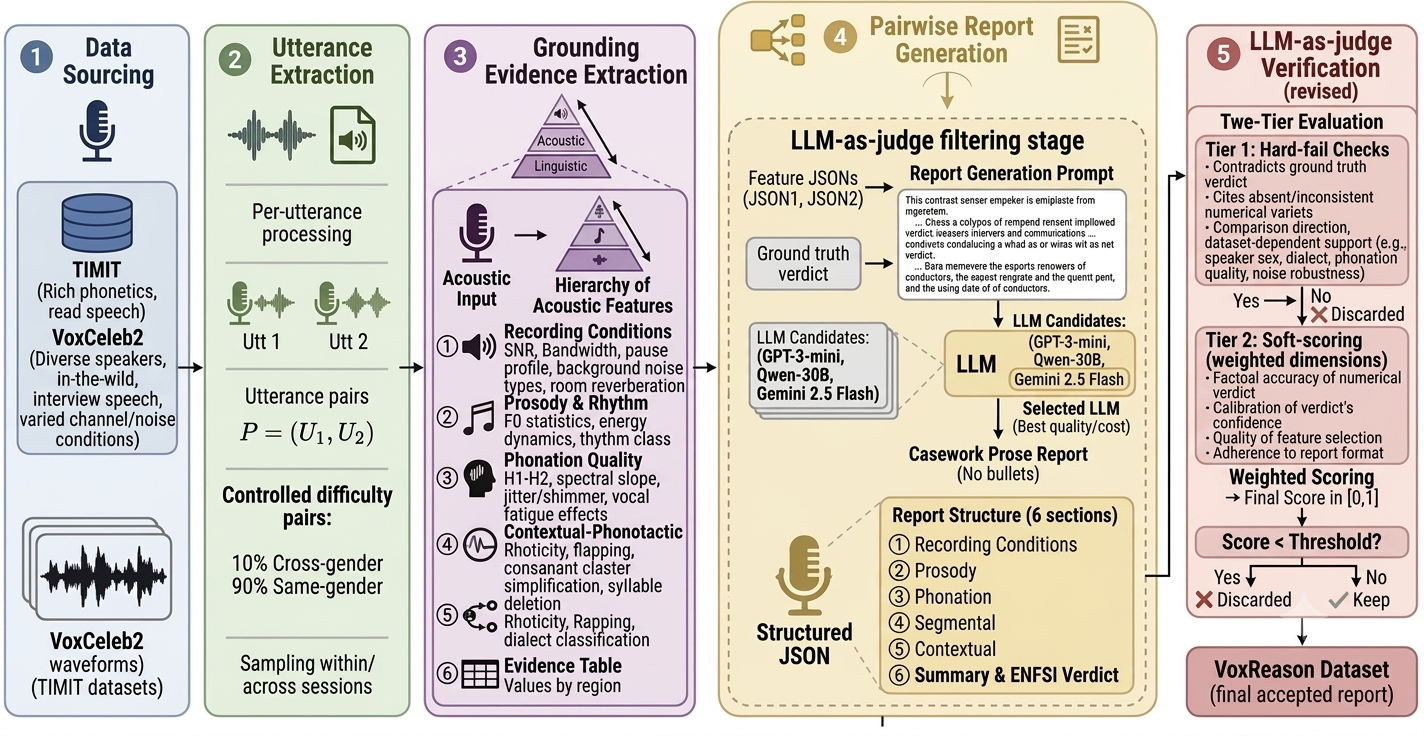}
    \caption{VoxReason construction: measurements are extracted from each recording, an LLM writes a comparison report from the two measurement records and the label, and a deterministic numerical validator and an LLM judge filter the reports.}
    \label{fig:data_pipeline}
\end{figure}

VoxReason is the dataset we build to train and evaluate CoLMbo-SV's
reports. Each example is a pair of recordings with four things attached:
the two utterances, their extracted acoustic measurements, the speaker
label, and a comparison report whose numerical claims have been checked
against those measurements (Figure~\ref{fig:data_pipeline}; details in
Appendix~\ref{app:voxreason}).

It exists because no existing corpus supplies this combination.
Verification corpora \citep{nagrani2017voxceleb} have speaker labels but no
language. Captioning corpora \citep{kim2019audiocaps,drossos2020clotho}
have language, but they describe single clips, not a comparison between
two, and their claims are not tied to measurements. A model that must write
a checkable comparison needs both halves in one example.

VoxReason holds 40{,}000 training pairs, half from TIMIT (462 speakers) and half from
VoxCeleb2 (700 speakers), balanced between same- and different-speaker
trials, and 1{,}000 test pairs per source from speakers disjoint from
training (Table~\ref{tab:voxreason_stats}). VoxReason is
supervision for comparative \emph{reporting}. It is not ground truth for
\emph{explaining} a verifier's decision. It fixes what the recordings
measure and validates language about those measurements. Whether a report
identifies the evidence behind the decision is a separate question, which
\S\ref{sec:tcav} answers.

\paragraph{Recordings and measurements.}
Pairs are drawn from TIMIT \citep{garofolo1993darpa}, phonetically rich read
speech across eight U.S.\ dialect regions, and VoxCeleb2
\citep{chung2018voxceleb2}, a large speaker population recorded in the wild
under varied channel and noise conditions (pair construction:
Appendix~\ref{app:voxreason}). For every utterance we extract acoustic measurements
spanning recording conditions (SNR, noise floor, speech ratio), prosody (F0
mean, SD and range, voicing ratio, syllable rate, energy dynamics), phonation
(jitter, shimmer, HNR, breathiness, perceptual voice-quality descriptors),
and segmental acoustics (F1--F3 means, voiced and voiceless VOT). Eighteen of
these scalars are the \emph{measurements} referred to throughout: the
evidence a report can cite and the quantities its claims are checked against.

\paragraph{Comparison reports.}
For each pair we prompt an LLM (Gemini~2.5~Flash-Lite, selected on a
100-pair development set) with the two measurement records and the speaker
label to produce a comparison report in continuous casework prose, in a
structured format motivated by forensic voice-comparison reporting
(recording conditions, prosody, phonation, segmental acoustics, weight of
evidence), terminated by a same/different conclusion, with numerical values
embedded in analytical sentences, and a closing statistical-strength
sentence that is part of the format, not an evaluated output
(Appendix~\ref{app:voxreason}).

\paragraph{Filtering.}
Given the label in the prompt, LLM generators rationalise: they retrofit a
plausible justification to the given verdict and cite values that do not
appear in the measurement records \citep{turpin2023language,lyu2023faithful}.
Because in this task the numerical claims \emph{are} the evidence, the
primary safeguard is deterministic and does not depend on any model: a
validator extracts every numerical literal, discards non-claim matches
(section numbers, formant labels, utterance indices, statistic artefacts),
and checks each remaining number against the union of the two measurement
records within an absolute tolerance of 0.5 or a relative tolerance of 5\%,
rejecting a report if more than 20\% of its claims fail; 91.9\% of
training generations have every cited number verified and 0.04\% exceed the
threshold (Appendix~\ref{app:voxreason}). An LLM-as-judge
\citep{liu2023g,zheng2023judging}, given the records, the label and the
report, then screens only for errors numerical matching cannot see: a
verdict contradicting the label, an inverted A-versus-B comparison,
unsupported categorical claims about dialect, sex or phonation, and a
weighted soft score over accuracy, calibration, feature selection and
format below which a report is discarded.

%% file: sections/eval.tex
\label{sec:eval}

The introduction's three questions become three measurements: what the
report \emph{describes} (numerical grounding, under a protocol that copying
cannot satisfy), what the decision \emph{depends on} (sensitivity of each
decision path to acoustic concept directions), and what the report
\emph{discusses} (decision grounding: are the selected features the ones
the decision is sensitive to).

\paragraph{Verification.}
Every EER is computed from a \emph{verdict score}: for CoLMbo-SV and the
audio-language baselines, the difference in teacher-forced negative
log-likelihood between the two candidate conclusions (``Different speakers''
minus ``Same speaker''), swept over every unique score as a threshold; for
dedicated encoders, cosine similarity. VoxCeleb1-O (37{,}611 trials) and
SITW core--core (721{,}788 trials) are used at full scale.

\paragraph{Numerical grounding.}
We run the VoxReason extractor over every VoxCeleb1-O recording
(4{,}708 files), build a 3{,}000-pair evaluation set in the VoxReason
schema, and score a stratified sample of $N{=}600$ pairs for every system on
identical pair identities. The \emph{numerical grounding} of a report is
the fraction of its numerical claims that match a measurement within the
validator's tolerance; \emph{perfectly grounded} and $\geq\!50\%$
\emph{grounded} are the fractions of reports in which all, respectively at
least half, of the cited numbers are verifiable. All three are computed over
reports that cite at least one number. \emph{Feature coverage} is the
fraction of the features the prompt asks about that the report addresses;
\emph{repetition} is the mean fraction of repeated 3-grams. A report that
never commits to a verdict is \emph{unknown}, and wrong under \emph{strict}
accuracy.

\paragraph{Numbers are withheld from the prompt.}
An audio-language model whose prompt contains the measurement block can reach
high grounding by copying. Every baseline prompt is therefore the bare
section instruction, verified digit-free; the measurement-in-prompt condition
is reported once, for fine-tuned Qwen2-Audio, as a copying upper bound.
CoLMbo-SV's measurement channel is not a prompt in this sense: the values are
computed from the audio by the system itself (\S\ref{sec:method}).

\paragraph{Does the report explain the decision?}
Three steps on the same 600 pairs. \emph{(i)~What the embedding encodes.}
For each of the eighteen measurements and two voice-quality descriptors we
fit a linear probe in the joint space
$z=[\hat{\mathbf{e}}_a;\hat{\mathbf{e}}_b]$ separating pairs in the top
tercile of the signed gap $x_a-x_b$ from the bottom tercile, on 20{,}000
VoxCeleb2 pairs from 700 speakers disjoint from the test speakers, with
speaker-disjoint folds. A concept is \emph{encoded} if held-out balanced accuracy is $\geq 0.70$ with a
200-permutation $p<0.01$; its unit weight vector is the concept direction
$v_c$. \emph{(ii)~What the decision is sensitive to.} We differentiate each
decision path with respect to $z$ --- the head's logit and the language
model's verdict score, back-propagated through mapper and language model
--- antisymmetrise against the swapped pair, project $v_c$ onto the tangent
space, and report the magnitude share
$M_c=\mathbb{E}|\nabla_z h\cdot v_c|/\mathbb{E}\|\nabla_z h\|$
\citep{kim2018tcav}, tested by rank against 200 random directions with
Benjamini--Hochberg correction \citep{benjamini1995fdr}; the sign-based TCAV
score is degenerate here (Appendix~\ref{app:tcav}). This is a sensitivity
analysis, not an intervention: the \emph{decision-relevant set} $C^*$ is the
concepts both encoded and significant. \emph{(iii)~What the report
discusses.} \emph{Decision grounding} (DG) is the fraction of the features a
report discusses that are in $C^*$; \emph{DG@3} is the fraction of the
pair's own three highest-$M$ concepts it mentions, and its \emph{lift} is
DG@3 minus the same quantity against a random other pair's top three, which
separates a report that responds to the pair from one that recites a fixed
list. Finally we test whether the gap is a matter of instruction by naming
each pair's top-three concepts in the prompt and, as a control, another
pair's.

\paragraph{Baselines.}
Encoders ReDimNet-B6 \citep{yakovlev2024redimnet}, ECAPA-TDNN
\citep{desplanques2020ecapa} and our frozen W2V-BERT~2.0; audio-language
models Qwen2-Audio-7B \citep{chu2024qwen2audio}, zero-shot and
LoRA-fine-tuned on VoxReason, and Kimi-Audio-7B \citep{team2025kimi},
LoRA-fine-tuned on VoxReason.

%% file: sections/results.tex
\label{sec:results}

\begin{table}[t]
\centering
\small
\setlength{\tabcolsep}{6pt}
\renewcommand{\arraystretch}{1.0}
\caption{Verification EER (\%) on VoxCeleb1-O and SITW, full scale; audio
LMs fine-tuned on VoxReason unless marked zero-shot.}
\label{tab:eer}
\begin{tabular}{@{}llrr@{}}
\toprule
System & Report & Vox1-O & SITW \\
\midrule
\multicolumn{4}{@{}l}{\emph{Dedicated speaker encoders (cosine)}} \\
ReDimNet-B6                     & no  & \textbf{0.17} & \textbf{0.28} \\
W2V-BERT~2.0 (our encoder)      & no  & 0.34 & 0.87 \\
ECAPA-TDNN                      & no  & 0.90 & 1.45 \\
\midrule
\multicolumn{4}{@{}l}{\emph{Audio-language models}} \\
Qwen2-Audio-7B, zero-shot        & yes & 47.01 & 47.67 \\
Qwen2-Audio-7B, fine-tuned       & yes & 35.08 & 20.97 \\
Kimi-Audio-7B, fine-tuned        & yes &  4.83 & 5.87 \\
\midrule
CoLMbo-SV (ours)                 & yes & \textbf{0.99} & \textbf{3.39}  \\
\bottomrule
\end{tabular}
\end{table}

\paragraph{Can it verify?}
Table~\ref{tab:eer} places CoLMbo-SV between the two families it draws on.
It reaches 0.99\% EER on VoxCeleb1-O while writing a report for every
trial: 79.5\% lower error than Kimi-Audio (4.83\%) and 97.2\% lower than
Qwen2-Audio (35.08\%), the two audio-language models fine-tuned on the same
data. This closes most of the gap between an audio-language model and
dedicated speaker encoders (ECAPA-TDNN 0.90\%; the frozen W2V-BERT encoder
inside CoLMbo-SV 0.34\%) while retaining report generation; the remaining
gap is the price of routing the decision through a language model that
must also write (\S\ref{sec:ablation}). Zero-shot Qwen2-Audio is at chance
(47.01\%): pairwise comparison is not a pretrained capability. The
ordering holds on SITW, out of domain for every model: 3.39\% against
5.95\% for Kimi-Audio.

\begin{table}[t]
\centering
\footnotesize
\setlength{\tabcolsep}{3pt}
\renewcommand{\arraystretch}{1.02}
\caption{Report quality on VoxCeleb1-O, $N{=}600$ identical pairs, no
numerical values in any baseline prompt.}
\label{tab:explain}
\setlength{\tabcolsep}{2.5pt}
\begin{tabular}{@{}lrrrrrrrr@{}}
\toprule
 & \multicolumn{3}{c}{Verdict (\%)} & \multicolumn{3}{c}{Numerical grounding} & & \\
\cmidrule(lr){2-4}\cmidrule(lr){5-7}
System & Strict$\uparrow$ & Answ.$\uparrow$ & Unk.$\downarrow$
       & Score$\uparrow$ & Perf.$\uparrow$ & $\geq$50\%$\uparrow$ & Cov.$\uparrow$ & Rep.$\downarrow$ \\
\midrule
Qwen2-Audio, zero-shot        &  3.83 & 46.94 & 91.83 & 0.068 & 17.5  & 56.7 & 25.7 & 0.035 \\
Qwen2-Audio, fine-tuned       & 89.33 & 90.39 & 1.17 & 0.671 & \textbf{16.2} & 80.8 & 77.6 & \textbf{0.015} \\
CoLMbo-SV (ours)              & \textbf{94.00} & \textbf{94.16} & \textbf{0.17} & \textbf{0.820} & 0.2 & \textbf{100.0} & \textbf{94.7} & 0.074 \\
\midrule
Qwen2-Audio, FT, measurements in prompt$^{*}$ & 24.00 & 84.71 & 71.67 & 0.734 & 19.3 & 99.8 & 67.8 & 0.028 \\
\bottomrule
\end{tabular}
\end{table}

\paragraph{Can it report grounded numbers without being shown them?}
Table~\ref{tab:explain} answers the first of the three questions under the
protocol of \S\ref{sec:eval}: no baseline prompt contains a number, so a
grounded claim must come from the audio, or, for CoLMbo-SV, from
measurements the system extracted itself. Under it, 82\% of the numerical
claims in CoLMbo-SV's reports match the measurements, every report is at
least half grounded, and 0.17\% fail to reach a verdict. The 0.820 score is
claim-level and should not be read as 82\% of reports being entirely
correct: only 0.2\% satisfy the all-claims criterion, because CoLMbo-SV
cites about 100 numbers per report and a single miss zeroes the indicator.
Fine-tuned Qwen2-Audio grounds 67\% of its claims and wins the
perfectly-grounded rate (16.6\%) and repetition, both effects of citing
about 16 numbers in short reports; zero-shot Qwen2-Audio does not answer
(91.83\% unknown). The copying row shows why the protocol matters: with the measurements in
its prompt Qwen2-Audio reaches 0.823 but stops answering (71.67\%
unknown). These numbers
establish the first half of the paper's claim; whether the evidence the
report describes is the evidence the decision uses is
\S\ref{sec:tcav}'s question (in-domain TIMIT results against
Qwen3-Omni-30B \citep{xu2025qwen3}: Appendix~\ref{app:timit}).

%% file: sections/ablation.tex
\label{sec:ablation}

\begin{table}[t]
\centering
\footnotesize
\setlength{\tabcolsep}{5pt}
\renewcommand{\arraystretch}{1.04}
\caption{Ablation. B is fine-tuned from A; C, D and E are each fine-tuned
from B with the change named; E is CoLMbo-SV. Verdict EER on VoxCeleb1-O;
report metrics on the 600 pairs of Table~\ref{tab:explain}.}
\label{tab:ablation}
\begin{tabular}{@{}clrrrrr@{}}
\toprule
 & Variant & EER$\downarrow$ & Ground.$\uparrow$ & $\geq$50\%$\uparrow$ & Unk.$\downarrow$ & Rep.$\downarrow$ \\
\midrule
A & stage 1: verifier, no evidence block in targets
    & 1.06 & 0.639 & 89.0 & \textbf{0.00} & 0.081 \\
B & stage 2: report objective, verifier frozen
    & 0.78 & 0.631 & 88.2 & 0.17 & 0.077 \\
C & \ + evidence-block targets, no measurements given
    & 1.37 & 0.658 & 90.3 & \textbf{0.00} & \textbf{0.070} \\
D & \ + measurements as 8 soft tokens
    & \textbf{0.76} & 0.686 & 95.0 & 0.67 & \textbf{0.070} \\
E & \ + measurements as text (CoLMbo-SV, stage 3)
    & 0.99 & \textbf{0.820} & \textbf{100.0} & 0.17 & 0.074 \\
\bottomrule
\end{tabular}
\end{table}

Table~\ref{tab:ablation} isolates what each component of
\S\ref{sec:method} contributes; all variants share the frozen encoder, the
mapper with its bypass, the backbone and the data. Three findings follow.

\paragraph{Finding 1: the report objective helps verification only if the verifier is frozen.}
Continuing the stage-1 verifier on the report objective with the head frozen
(B) improves the written verdict from 1.06\% to 0.78\% EER: the language
model learns to write, and to read the comparison token, without the report
loss moving the decision it conditions on.

\paragraph{Finding 2: evidence-block targets cost verification unless the numbers are supplied.}
Asking the same model to open every report with a structured evidence block
while giving it no measurements (C) raises EER to 1.37\%: the verdict now
follows a hundred numbers the model must invent. Supplying the measurements
repairs it (D 0.76\%, E 0.99\%): what costs verification is not the block
but writing it blind.

\paragraph{Finding 3: soft tokens carry little of the numbers; text carries them.}
D and E receive the same eighteen measurements and differ only in delivery.
In D a two-layer MLP maps the standardised measurement vector to eight soft
tokens, with a reconstruction decoder at its floor, so the values are
recoverable from the tokens. Yet D's grounding, 0.686, is only 0.028 above C's, which is given
no measurements at all, whereas E's is 0.162 above: for ground-truth mean
$F_0$ of 146.4/356.1\,Hz, D writes 111.2/206.2\,Hz and E writes
146.4/356.1\,Hz. The language model reads few digits out of a learned
vector; it copies them from tokens. Text delivery raises grounding to 0.820
with every report at least half grounded, at a cost of 0.23 EER points as
the verdict now attends to a 400-token block. D is the better verifier; E
is the only configuration whose reports can be checked against the signal,
which is why E is CoLMbo-SV (Appendix~\ref{app:decoding}).

%% file: sections/tcav.tex
\label{sec:tcav}

Section~\ref{sec:results} shows that the reports describe the recordings
correctly; this section asks whether they describe the evidence the
decision uses (three steps, \S\ref{sec:eval}).

\paragraph{What the embedding encodes.}
Fourteen of twenty concepts pass the probe gate (Appendix~\ref{app:tcav}).
Mean $F_0$ is the most decodable (balanced accuracy 0.905), then breathiness
(0.878), roughness (0.875) and HNR (0.824). Voice-onset time is at chance
(0.56--0.57) and speaking rate barely above it (0.60), yet both are cited by
every model and feature coverage rewards it: a report can discuss a feature
that the speaker representation does not reliably encode at all. The verdict
itself is bilinear, not linear, in $z$ (cosine AUC 0.9996, linear probe
0.49; Appendix~\ref{app:tcav}).

\paragraph{What the decision is sensitive to.}
The two decision paths of \S\ref{sec:method} attribute the verdict
differently. For the verification head only two concepts survive
correction, $F_0$ range and $F_0$ standard deviation, and pitch variability
is its top-ranked concept on 92\% of pairs. For the language model's verdict
fourteen survive --- voice quality (roughness, breathiness, shimmer, HNR,
jitter, flutter), recording conditions (SNR, noise floor, speech ratio),
voicing ratio, pitch (mean, range, SD) and $F_3$ --- and the ranking varies
from pair to pair (118 distinct top-three sets over 600 pairs). The
separation is by design: nothing constrains the continuous score and the
written conclusion to use the same evidence, so explaining one is not
explaining the other; mean $F_0$, the best-encoded concept, is not among
the head's concepts at all.

\begin{table}[t]
\centering
\footnotesize
\setlength{\tabcolsep}{4pt}
\renewcommand{\arraystretch}{1.04}
\caption{Decision grounding of the reports of Table~\ref{tab:explain}
against the language model's decision-relevant set ($|C^*|{=}13$ of 16
report features; random selection scores DG${}=0.812$). DG: share of
discussed features in $C^*$; Cov: share of $C^*$ covered; DG@3: share of
the pair's own top-3 discussed; shuf.: same against a random other pair;
lift: their difference.}
\label{tab:cg}
\begin{tabular}{@{}lrrrrrr@{}}
\toprule
System & feats & DG & Cov & DG@3 & shuf. & lift \\
\midrule
CoLMbo-SV                                & 6.8 & 0.835 & 0.466 & 0.523 & 0.496 & 0.027 \\
\ \ + prompt names this pair's top-3      & 6.8 & 0.834 & 0.464 & 0.522 & 0.494 & 0.028 \\
\ \ + prompt names another pair's top-3   & 6.8 & 0.835 & 0.465 & 0.517 & 0.492 & 0.024 \\
Qwen2-Audio, fine-tuned                   & 4.8 & 0.830 & 0.320 & 0.443 & 0.351 & 0.092 \\
Qwen2-Audio, FT, measurements in prompt   & 6.9 & 0.936 & 0.487 & 0.534 & 0.512 & 0.023 \\
Qwen2-Audio, zero-shot                    & 1.0 & 0.917 & 0.071 & 0.181 & 0.172 & 0.010 \\
\bottomrule
\end{tabular}
\end{table}

\paragraph{What the report discusses.}
Table~\ref{tab:cg} scores the reports of Table~\ref{tab:explain} against
the language model's decision-relevant set. The pair-specific measure is
the informative one: of a pair's own three strongest concepts CoLMbo-SV
mentions 52.3\%, and of a random other pair's, 49.6\%, a lift of 0.027. The
report discusses this pair's evidence almost exactly as often as another
pair's. The aggregate hides this: 83.5\% of the 6.8 features a report
discusses are decision-relevant, but with thirteen of sixteen features in
$C^*$ random selection would score 81.2\%. Fine-tuned Qwen2-Audio is more
selective (lift 0.092) while covering a third less of $C^*$, and loses that
selectivity (0.023) once the measurements are in its prompt. Reports that score well on grounding and coverage thus show little
selectivity for the evidence behind the particular decision.

\paragraph{Prompting does not close it.}
The second and third rows of Table~\ref{tab:cg} regenerate
CoLMbo-SV's reports with each pair's own top-three concepts named in the
prompt, and with a random other pair's as a control; neither moves any
column. The named features appear 48.6\% of the time when they are the
pair's own, 47.3\% when they are another pair's, and 48.8\% with no
instruction, and 79\% of reports name exactly the feature set the
un-instructed model produces: a template learned from the training
targets, not the pair in front of it, decides what the report says. Making
reports selective therefore requires supervision that connects each report to the evidence
responsible for its particular decision, which VoxReason, by construction
(\S\ref{sec:dataset}), does not provide.

%% file: sections/appendix.tex
\section{Related work}
\label{app:related}
\input{sections/lit}

\section{VoxReason: construction details}
\label{app:voxreason}

\begin{table}[h]
\centering
\small
\setlength{\tabcolsep}{7pt}
\caption{VoxReason composition. Train and test speakers are disjoint within
each source. Every pair carries one validated report; the 761{,}151 training
rows are the reports and their per-section and per-feature decompositions
used as instruction targets.}
\label{tab:voxreason_stats}
\begin{tabular}{@{}llrrrr@{}}
\toprule
Split & Source & Pairs & Same & Different & Speakers \\
\midrule
Train & TIMIT     & 20{,}000 & 10{,}000 & 10{,}000 & 462 \\
Train & VoxCeleb2 & 20{,}000 & 10{,}000 & 10{,}000 & 700 \\
Test  & TIMIT     &  1{,}000 &     500 &     500 & 168 \\
Test  & VoxCeleb2 &  1{,}000 &      -- &      -- & 118 \\
\bottomrule
\end{tabular}
\end{table}

\paragraph{Pair construction and validation yield.}
Same-speaker pairs take two different utterances of one speaker.
Different-speaker pairs are stratified by sex and, for TIMIT, dialect
region (of TIMIT's 10{,}000 different-speaker training pairs, 5{,}000 share
region and sex, 2{,}000 share region only, 2{,}000 share sex only, 1{,}000
share neither); VoxCeleb2 negatives are stratified by sex. Reports were
generated with Gemini~2.5~Flash-Lite, one per pair, with the label and both
measurement records in the prompt. The deterministic validator finds every
cited number verifiable in 91.9\% of training generations (18{,}363 of
20{,}000 TIMIT; 18{,}396 of 20{,}000 VoxCeleb2) and in 92.2\% and 91.6\% of
the two test sets; across all 40{,}000 training generations the mean
unverified fraction is 0.4\%, sixteen generations (0.04\%) exceed the 20\%
rejection threshold, and none contradicts its label in the conclusion.

\paragraph{Statistical-strength sentence.}
Free-form LLM reports attach verbal confidence (``strongly suggests'',
``high confidence'') inconsistently. To replace these hedges with a
measurable quantity in the supervision, each report closes with a sentence
computed from the measurements. For each discriminative scalar feature $f$
we precompute a sex-conditional within-speaker standard deviation
$\sigma_w(f,g)$ from training-side speakers and score the pair with
$z_f=|v_a^{(f)}-v_b^{(f)}|/\sigma_w(f,g)$, $p_f=2(1-\Phi(z_f))$; the
per-feature $p$-values are combined by Fisher's method,
$\chi^2=-2\sum_f\ln p_f\sim\chi^2_{2k}$, and discretised into a four-level
verbal-scale label ($p>0.40$ same/high; $0.25<p\le0.40$ same/medium;
$0.05<p\le0.25$ different/medium; $p\le0.05$ different/high).
Recording-condition features are excluded from the combination. The sentence
reports the top three discriminating features and their $z$-statistics. It is
part of the report format the model learns to write; its calibration is not
evaluated in this paper, and the verdict metrics parse only the same/different
conclusion. Representative outputs of the TIMIT-trained checkpoint, including
this sentence, are shown in Table~\ref{tab:examples}.

\section{In-domain results on TIMIT}
\label{app:timit}

\begin{table}[h]
\centering
\small
\setlength{\tabcolsep}{6pt}
\caption{In-domain comparison against zero-shot Qwen3-Omni-30B-A3B-Thinking
on the TIMIT pairs test split ($N{=}1917$ matched pairs). Repetition here is
the fraction of reports with any repeated 3-gram. Best in bold.}
\label{tab:main_results}
\begin{tabular}{@{}lrr@{}}
\toprule
                                & Qwen3-Omni & CoLMbo-SV \\
\midrule
Strict verdict (\%) $\uparrow$           & 24.31 & \textbf{77.10} \\
Answered verdict (\%) $\uparrow$         & 48.54 & \textbf{77.10} \\
Unknown rate (\%) $\downarrow$           & 49.92 & \textbf{ 0.00} \\
\midrule
Numerical grounding $\uparrow$           &  0.23 & \textbf{ 0.70} \\
Perfectly grounded (\%) $\uparrow$       &  9.12 & \textbf{47.47} \\
$\geq 50\%$ grounded (\%) $\uparrow$     & 23.20 & \textbf{84.42} \\
Feature coverage (\%) $\uparrow$         & 71.05 & \textbf{95.36} \\
Repetition (\%) $\downarrow$             & 92.07 & \textbf{ 1.62} \\
\bottomrule
\end{tabular}
\end{table}

\begin{table}[h]
\centering
\small
\caption{Verdict accuracy and numerical grounding on TIMIT pairs stratified
by minimum SNR ($N{=}3000$). Best in bold. $^{\dagger}$Sample size is small;
read as suggestive.}
\label{tab:snr_strat}
\setlength{\tabcolsep}{8pt}
\begin{tabular}{lrrrrr}
\toprule
SNR bucket          & $n$    & Mean SNR (dB) & Verdict (\%) & Grounding & Perfect (\%) \\
\midrule
$\le 15$ dB$^{\dagger}$ &     4  & 13.06         & 25.00          & 0.44         &  0.00       \\
$15$–$25$ dB         &   142  & 21.75         & 71.83          & 0.71         & 48.23       \\
$25$–$35$ dB         &  1672  & 30.96         & 73.15          & 0.73         & 48.85       \\
$> 35$ dB            &  1182  & 37.83         & \textbf{79.02} & 0.72         & 44.59       \\
\bottomrule
\end{tabular}
\end{table}

Stratifying by the minimum SNR within each pair (Table~\ref{tab:snr_strat})
shows a monotone relationship between recording quality and verdict
accuracy, from 25\% below 15\,dB to 79.0\% above 35\,dB, with grounding
moving the same way (0.44 to 0.72); the same trend holds for noise floor
(69.4\% $\rightarrow$ 78.0\%) and speech ratio (64.5\% $\rightarrow$ 79.6\%).
Because the model emits these measurements as the first section of every
report, the user receives both a decision and a record of the conditions
under which it was made.

\section{Prompt-conditioned routing in the mapper}
\label{app:routing}

\begin{figure}[h]
    \centering
    \includegraphics[width=0.95\textwidth]{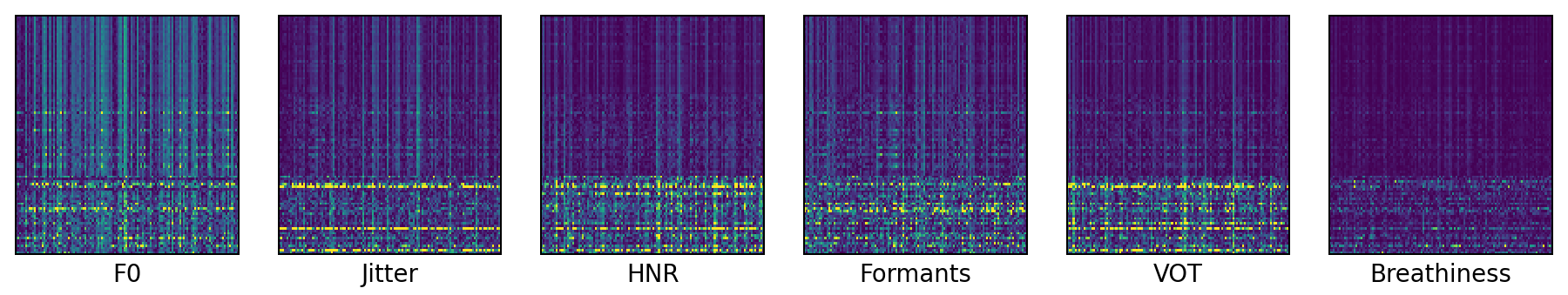}
    \caption{Absolute prefix shift $|h_{\text{probe}}-h_{\text{neutral}}|$
    induced by six feature-specific prompts relative to the neutral pairwise
    prompt, over the 120 most variable prefix dimensions (columns) and the
    prefix positions (rows), averaged over 100 test pairs.}
    \label{fig:probe_maps}
\end{figure}

\begin{table}[h]
\centering
\small
\setlength{\tabcolsep}{8pt}
\caption{Mapper prefix shift from the neutral baseline under feature-specific
prompts (100 pairs), alongside each feature's decision-sensitivity magnitude $M$ for the
language-model verdict (best of its constituent measurements).}
\label{tab:probe_diff}
\begin{tabular}{@{}lrrr@{}}
\toprule
Probe       & $\Delta_F$ & Cos.\ dist. & $M$ (sensitivity) \\
\midrule
Breathiness & 0.146 & 0.010 & 0.140 \\
F0          & 0.180 & 0.015 & 0.098 \\
Jitter      & 0.189 & 0.017 & 0.116 \\
HNR         & 0.197 & 0.018 & 0.122 \\
Formants    & 0.228 & 0.025 & 0.079 \\
VOT         & 0.228 & 0.024 & 0.059 \\
\bottomrule
\end{tabular}
\end{table}

A design goal of the mapper is that a query about a specific feature draw on
the corresponding part of the speaker representation rather than a
prompt-invariant summary. Running the mapper on 100 test pairs under six
feature-specific prompts and a neutral prompt, the relative Frobenius shift of
the prefix ranges from 14.6\% (breathiness) to 22.8\% (formants, VOT), a 42\%
spread (Table~\ref{tab:probe_diff}, Figure~\ref{fig:probe_maps}): the mapper
does produce a query-dependent prefix. The shift does not track the verdict's
sensitivity, however. The two queries that move the prefix most name the
features with the smallest sensitivity magnitude, and the query that moves it least
names one of the largest; with six probes the rank correlation is negative but
not significant, so we state this only as an observation. A large
prompt-induced shift shows that the mapper re-encodes the query, not that the
verdict uses the queried feature --- the same distinction between presence and
use that \S\ref{sec:tcav} measures directly.

\section{Decoding ablation}
\label{app:decoding}

\begin{table}[h]
\centering
\small
\setlength{\tabcolsep}{7pt}
\caption{Decoding controls (variant D of Table~\ref{tab:ablation}, first 200
evaluation pairs).}
\label{tab:decoding}
\begin{tabular}{@{}rrrrr@{}}
\toprule
Rep.\ penalty & No-repeat $n$-gram & Grounding$\uparrow$ & Rep.\ rate$\downarrow$ & Unk.\ (\%)$\downarrow$ \\
\midrule
1.00 & --  & \textbf{0.740} & 0.277 & 6.0 \\
1.15 & --  & 0.728 & 0.127 & 8.0 \\
1.15 & 6   & 0.696 & 0.070 & 1.5 \\
1.15 & 4   & 0.678 & \textbf{0.036} & \textbf{0.5} \\
\bottomrule
\end{tabular}
\end{table}

Greedy decoding with no repetition control produces looping in long reports.
A repetition penalty with $n$-gram blocking cuts the 3-gram repetition rate
from 0.277 to 0.036 and the unknown rate from 6.0\% to 0.5\% --- blocked loops
no longer consume the budget before the conclusion --- but each step of
blocking also lowers grounding, from 0.740 to 0.678, because blocked
$n$-grams include the repeated feature-name phrasing the evidence format
legitimately reuses. All reported CoLMbo-SV numbers use penalty 1.15 with
6-gram blocking.

\paragraph{A from-scratch variant.}
A variant that also moved the comparison and measurement tokens out of the
cross-projection changes the prefix layout and could not be resumed from
stage~2; trained from scratch for the same budget it reached chance EER
despite a perfect head and a reconstruction loss at its floor. The
information was in the prefix and the language model never learned to read
it: the presence-versus-use gap that \S\ref{sec:tcav} measures.

\section{Concept probes and sensitivity: full table}
\label{app:tcav}

\begin{table}[h]
\centering
\footnotesize
\setlength{\tabcolsep}{3pt}
\caption{Which acoustic concepts the frozen speaker embedding encodes, and
which of them the verdict is sensitive to. \emph{Probe} is speaker-disjoint
held-out balanced accuracy (gate $\geq 0.70$); $M$ and $p$ are the magnitude
share and its empirical $p$-value against 200 random directions for the
language-model verdict of CoLMbo-SV. The last two columns mark concepts that
are both encoded and significant after BH correction, for the language-model
verdict and for the verification head. $^{a}$significant but not encoded, so
the direction is not interpretable. Rows sorted by $M$.}
\label{tab:tcav}
\resizebox{0.7\linewidth}{!}{%
\begin{tabular}{@{}lrcrrcc@{}}
\toprule
 & & & & & \multicolumn{2}{c}{Decision-relevant} \\
\cmidrule(lr){6-7}
Concept & Probe & Enc. & $M$ (LM) & $p$ & LM & Head \\
\midrule
roughness (VQ)     & 0.875 & \checkmark & 0.142 & $<$0.005 & \checkmark & -- \\
breathiness        & 0.878 & \checkmark & 0.140 & $<$0.005 & \checkmark & -- \\
shimmer            & 0.786 & \checkmark & 0.125 & $<$0.005 & \checkmark & -- \\
voicing ratio      & 0.778 & \checkmark & 0.124 & $<$0.005 & \checkmark & -- \\
HNR                & 0.824 & \checkmark & 0.122 & $<$0.005 & \checkmark & -- \\
jitter             & 0.762 & \checkmark & 0.116 & $<$0.005 & \checkmark & -- \\
energy SD          & 0.673 & --         & 0.107 & $<$0.005 & --$^{a}$   & -- \\
mean $F_0$         & \textbf{0.905} & \checkmark & 0.098 & $<$0.005 & \checkmark & -- \\
SNR                & 0.800 & \checkmark & 0.098 & $<$0.005 & \checkmark & -- \\
speech ratio       & 0.765 & \checkmark & 0.096 & $<$0.005 & \checkmark & -- \\
noise floor        & 0.756 & \checkmark & 0.090 & $<$0.005 & \checkmark & -- \\
$F_3$              & 0.703 & \checkmark & 0.079 & 0.005    & \checkmark & -- \\
$F_0$ range        & 0.764 & \checkmark & 0.076 & 0.010    & \checkmark & \checkmark \\
$F_0$ SD           & 0.755 & \checkmark & 0.076 & 0.010    & \checkmark & \checkmark \\
speaking rate      & 0.602 & --         & 0.075 & 0.010    & --$^{a}$   & -- \\
$F_1$              & 0.681 & --         & 0.068 & 0.020    & --$^{a}$   & -- \\
$F_2$              & 0.698 & --         & 0.065 & 0.020    & --$^{a}$   & -- \\
flutter (VQ)       & 0.735 & \checkmark & 0.065 & 0.020    & \checkmark & -- \\
voiceless VOT      & 0.561 & --         & 0.059 & 0.040    & --$^{a}$   & -- \\
voiced VOT         & 0.571 & --         & 0.053 & 0.085    & --         & -- \\
\bottomrule
\end{tabular}%
}
\end{table}

\paragraph{Why the sign-based TCAV score is not used.}
Kim et al.'s score counts the fraction of examples with a positive directional
derivative, which is informative only if the gradient direction varies across
examples. For the verification head the antisymmetrised gradient field has
mean cosine 0.94 with its own mean direction, so the sign of the derivative is
fixed by the concept alone: the random-direction null is bimodal (25\% of
random directions score below 0.05, 30\% above 0.95, sd 0.42) and a test
against it passes nearly everything. The magnitude share has a tight
unimodal null (mean 0.043, sd 0.029) and is used throughout. For the
language-model verdict the concentration is 0.41, and CG@3 is admissible
because the top-three cause set varies across pairs (118 distinct sets, the
modal one covering 10\% of pairs).

\paragraph{Limits of the analysis.}
The concept sets are not stratified by speaker sex, so the pitch and formant
directions may partly encode sex; the decision-relevant set is derived from CoLMbo-SV's
own language-model verdict and applied to every system's reports, which for
the baselines measures whether they discuss what \emph{our} verdict is
sensitive to; and feature mentions are lexical, so a feature counts as
discussed whether or not a number follows it.

\section{Representative outputs}
\label{app:examples}

\begin{table}[h]
\caption{Representative CoLMbo-SV outputs across prompt types, from the
TIMIT-trained checkpoint of Appendix~\ref{app:timit}. Numerical citations
match the measurements (grounding $\geq 0.9$ in each case). The closing
statistical-strength sentence and confidence label are part of the VoxReason
report format (Appendix~\ref{app:voxreason}) and are not evaluated.}
\label{tab:examples}
\begin{center}
\small
\setlength{\tabcolsep}{4pt}
\begin{tabular}{p{0.28\textwidth} p{0.64\textwidth}}
\toprule
\textbf{Prompt} & \textbf{Generated response} \\
\midrule
\textbf{Overall assessment.}
\textit{``Give an overall assessment of whether these two recordings come from the same speaker.''}
&
\textbf{\#\# 1. Recording Conditions.} \textit{``Utterance 1 exhibits a higher signal-to-noise ratio (SNR) of 35.49 dB compared to utterance 2's 31.21 dB. \dots\ While utterance 1 benefits from a cleaner recording environment, the differences in SNR and noise floor are not so substantial as to preclude a reliable acoustic comparison.''}

\textbf{\#\# 2. Prosodic \& Voice Quality.} \textit{``Mean F0: 113.24 vs 116.56 Hz; jitter higher in utterance 1 (1.99 vs 1.69); HNR 10.50 vs 11.90 dB.''}

\textbf{\#\# 3. Spectral.} \textit{``Utterance 1 F1=800 Hz, F2=1800 Hz, F3=2830 Hz; Utterance 2 F1=750, F2=1750, F3=2750.''} \\
\midrule
\textbf{Segmental analysis.}
\textit{``Analyze the segmental acoustic features F1, F2, F3 means, voiced and voiceless VOT and finish with a Conclusion line on speaker identity.''}
&
\textit{``Utterance 2 mean F1 827.08 Hz vs utterance 1's 756.44 Hz; utterance 1 F2=1859 Hz, F3=2919 Hz vs utterance 2's F2=1759, F3=2811. Voiceless VOT 80 ms (utt 2) vs 70 ms (utt 1); voiced VOT $-15$ vs $-18$ ms.''}

\textit{``Most discriminating evidence: F1 (z=1.16, p=0.245); F3 (z=0.78, p=0.437); F2 (z=0.80, p=0.424). Combining 5 features via Fisher's method ($\chi^2$=7.98, df=10, p=0.635) yields same speaker, high confidence.''}

\textbf{Conclusion:} \textit{Same speaker, high confidence.} \\
\midrule
\textbf{Single-feature query (HNR).}
\textit{``Examine HNR differences and explain how vocal clarity contributes to speaker comparison.''}
&
\textit{``Utterance 1 has higher HNR of 12.95 dB compared to utterance 2's 10.54 dB, indicating cleaner vocal tone.''}

\textit{``Typical within-speaker variation for HNR is 1.50 (utterance 1 12.95 dB, utterance 2 10.54 dB; z=1.64, p=0.100), indicating different speaker, medium confidence.''}

\textbf{Conclusion:} \textit{Different speakers, medium confidence.} \\
\midrule
\textbf{Single-feature query (breathiness).}
\textit{``Analyze breathiness in both utterances and explain whether it supports or weakens speaker differentiation.''}
&
\textit{``Breathiness index slightly higher in utterance 2 (0.55) vs utterance 1 (0.51).''}

\textit{``Most discriminating evidence: breathiness index (z=0.52, p=0.605); voice\_quality.breathiness (z=0.52, p=0.605); voice\_quality.whisper (z=0.12, p=0.903). Combining 3 features via Fisher's method ($\chi^2$=2.33, df=6, p=0.886) yields same speaker, high confidence.''}

\textbf{Conclusion:} \textit{Same speaker, high confidence.} \\
\bottomrule
\end{tabular}
\end{center}
\end{table}

%% file: sections/lit.tex
Forensic voice comparison concerns the comparison of known-speaker and questioned-speaker recordings to assist a court or other decision-maker in evaluating speaker identity. \citet{morrison2017admissibility} distinguish between the technical approach used to extract information from speech, the concrete system used to process a case, and the reasoning framework used to report the strength of evidence. Under the likelihood-ratio framework, the practitioner evaluates the probability of the observations under competing hypotheses, i.e. whether the questioned recording was produced by the known speaker or by some other speaker from a relevant population subgroup, which requires consideration of both similarity to the known speaker and typicality with respect to the population subgroup. Hence, similarity alone is not a sufficient strength-of-evidence statement. The ENFSI guidelines \citep{drygajlo2015guidelines} similarly frame automatic and semiautomatic forensic speaker recognition as an evaluative task embedded within a Bayesian interpretation framework, focusing on case assessment, net duration of speech, technical quality, mismatched recording conditions, relevant population, validation, calibration, and reporting.

Modern speaker verification systems represent each utterance as a dense speaker embedding and compare embeddings using cosine similarity, PLDA, or a learned backend. The x-vector family popularized neural embedding-based speaker recognition \citep{snyder2018xvectors}, and ECAPA-TDNN, with its channel-attentive temporal aggregation, remains a strong supervised baseline \citep{desplanques2020ecapa}. Self-supervised speech encoders such as wav2vec~2.0, HuBERT, WavLM, and W2V-BERT have further improved speaker representations when adapted to speaker-discriminative objectives \citep{chen2022wavlm}, achieving EERs below 1\% in VoxCeleb1. These systems are accurate but interpretable only in aggregate: the speaker embedding entangles pitch, voice quality, formant structure, segmental timing, and channel conditions into a single vector, and no per-feature reasoning can be recovered from a similarity score.

Recent forensic automatic speaker recognition research has emphasized that global performance metrics can hide substantial variability between speakers and samples. \citet{hughes2024individual} examine individual-speaker behavior in a forensic x-vector system and show that, despite strong overall performance, a minority of speakers produce substantially worse calibrated log-likelihood-ratio cost values. Their phonetic-content ablation further shows that vowels generally improve performance, while stops and fricatives can degrade it for many speakers, and that the contribution of phonetic classes varies by speaker. They also report that long-term laryngeal features, such as fundamental frequency and voice quality, are not well correlated with ASR performance, suggesting that such features may provide complementary information for problematic speakers. These findings motivate comparison-specific explanations that identify which phonetic and acoustic observations are informative, unreliable, or missing from the embedding-based decision.

A separate line of work seeks to make the speaker recognition score itself interpretable. The BA-LR framework represents each speech extract as a binary vector of shared speaker attributes, where each coefficient indicates the presence or absence of a latent voice attribute \citep{benamor2022balr}. Each attribute is associated with explicit behavioral parameters, and the final likelihood ratio is computed as a product of attribute-level likelihood ratios under an independence assumption. \citet{benamor2024bae} extend this approach with a binary autoencoder that natively produces binary attribute representations under an attribute-oriented loss that encourages dimensions to behave like shared speaker-specific attributes. This work is closely related to CoLMbo-SV in that it treats speaker recognition as decomposable evidence rather than opaque embedding comparison. CoLMbo-SV differs in that its evidence is grounded in named acoustic-phonetic measurements (F0, formants, jitter, shimmer, HNR, VOT, voice-quality descriptors) and is verbalized as a structured forensic-style report rather than a vector of latent binary attributes.

Phonetic-trait models offer another route toward interpretable speaker verification. ExPO \citep{ma2025expo} introduces phonetic trait layers into an ECAPA-TDNN-style architecture: a pretrained phone recognizer supplies phone boundaries, frame-level embeddings are averaged within phone segments, and the resulting phone-indexed trait embeddings are pooled into an utterance-level speaker embedding. At inference time, ExPO outputs both a final verification score and a phone-wise phonetic-trait similarity vector, allowing users to inspect which phonetic traits support a trial. This provides a human-interpretable view of neural speaker verification, but its explanations are learned phone-level similarity scores rather than named acoustic measurements.

Recent multimodal systems pair a speech encoder with a language model to perform audio captioning, speech understanding, and instruction following. Qwen-Audio and Qwen3-Omni \citep{chu2023qwenaudio} caption and answer questions about speech; SALMONN \citep{tang2024salmonn} integrates a Whisper encoder with an LLM via Q-Former layers; LTU and GAMA \citep{ghosh2024gama} train on instruction-following over audio content. These models can describe spectral and prosodic properties, but they are trained on caption-style data, not on pairwise speaker verification, and their numerical claims about a recording are typically not verified against the signal. In high-stakes speaker comparison, numerical claims must correspond to measured acoustic quantities, and the final decision must be traceable to the evidence cited in the report.

A small but growing body of work targets interpretable speech analysis with LLMs. The original CoLMbo system \citep{baali2025colmbo} produces natural-language descriptions of single utterances along dimensions such as accent, emotion, and voice quality, but it does not compare two recordings or render a verification decision. ADIFF \citep{ghosh2024adiff} performs a comparative voice-quality description for audio LLMs but operates only on perceived voice attributes; it does not extract acoustic measurements, and does not emit a verification verdict. Recent work on explainable speaker recognition has also explored attention-based saliency and prototype methods, but these explanations are post-hoc visualizations of a fixed-embedding decision rather than structural reformulations of the decision itself. CoLMbo-SV differs from these in producing a verification decision together with a report grounded in named acoustic measurements, and in evaluating that report not only for numerical grounding but for whether the evidence it discusses is the evidence the decision is sensitive to.